\documentclass[graybox]{svmult}

\usepackage{type1cm}        

\usepackage{makeidx}         
\usepackage{graphicx}        
\usepackage{multicol}        
\usepackage[bottom]{footmisc}
\usepackage{newtxtext}       %
\usepackage{newtxmath}       
\usepackage{natbib}

\usepackage{amssymb}
\usepackage{pifont}
\newcommand{\cmark}{\ding{51}}%
\newcommand{\xmark}{\ding{55}}%
\usepackage{blindtext}
\usepackage{hyperref}

\usepackage{fancyhdr}
\fancypagestyle{arxivhdr}
{
   \fancyhf{}
   \setlength{\headheight}{40pt} 
\fancyfoot[C]{This chapter appears in: Ang, M.H., Khatib, O., Siciliano, B. (eds) Encyclopedia of Robotics. Springer, Berlin, Heidelberg}
\fancyhead[C]{\footnotesize Please cite this chapter as:\\
Andreasson, H., Grisetti, G., Stoyanov, T., Pretto, A. (2023). Software Architectures for Mobile Robots. In: Ang, M.H., Khatib, O., Siciliano, B. (eds) Encyclopedia of Robotics. Springer, Berlin, Heidelberg. \url{https://doi.org/10.1007/978-3-642-41610-1\_160-1}}
}

\makeindex             

\begin{document}

\title*{Software Architectures for Mobile Robots}
\author{Henrik Andreasson, Giorgio Grisetti, Todor Stoyanov, and Alberto Pretto}
\institute{Andreasson and Stoyanov \at School of Science and Technology, \"{O}rebro University, S-701 82 \"{O}rebro, Sweden, \email{henrik.andreasson@oru.se, todor.stoyanov@oru.se}
\and Grisetti \at Department of Computer, Control, and Management Engineering ``Antonio Ruberti``, Sapienza University of Rome, Italy \email{grisetti@diag.uniroma1.it}
\and Pretto \at Department of Information Engineering, University of Padova, Italy \email{alberto.pretto@dei.unipd.it}}
%
%
\maketitle



\thispagestyle{arxivhdr}
\vspace{-2.0cm}
\section{Synonyms}

Middleware, software framework, system architecture.

\section{Definitions}

Software architecture, in general, both refers to the high-level structure of a system as well as to the process of ensuring that the structure or the design of a system is according to specific needs. For mobile robotics, specific requirements are, for example, real-time capabilities, asynchronous data processing, and distributed functionality.
While there is a clear distinction between a design of a software architecture suitable for robotics and the particular reference design implementation, in practice, due to the complexity of the task, frameworks for robotics often come with a single reference implementation.
Therefore, when comparing and choosing an appropriate software architecture, it is prudent to take into consideration not only the design but the suitability of the implementation as well. 

\section{Overview}

In the early stage of mobile robotics research, essentially every research group, or even individual researcher, had their own middleware solution to interface sensors and actuators, to log and visualize data, and so on. For a researcher the design and implementation of such system is usually a ``necessary evil'', as it is required in order to deploy subsequently developed research code. Only with respect to data logging, a plethora of different formats for storing sensory data have been proposed and used by the community, each necessitating its own set of data parsing tools and interfaces to convert to alternative formats. Optimal design of architectures suitable to the needs of a mobile robot system is a research topic on its own right, but the vast majority of researchers in the field are typically users of the middleware system, instead of active developers.

\begin{figure}[ht!]
   \centering
   \includegraphics[width=\columnwidth]{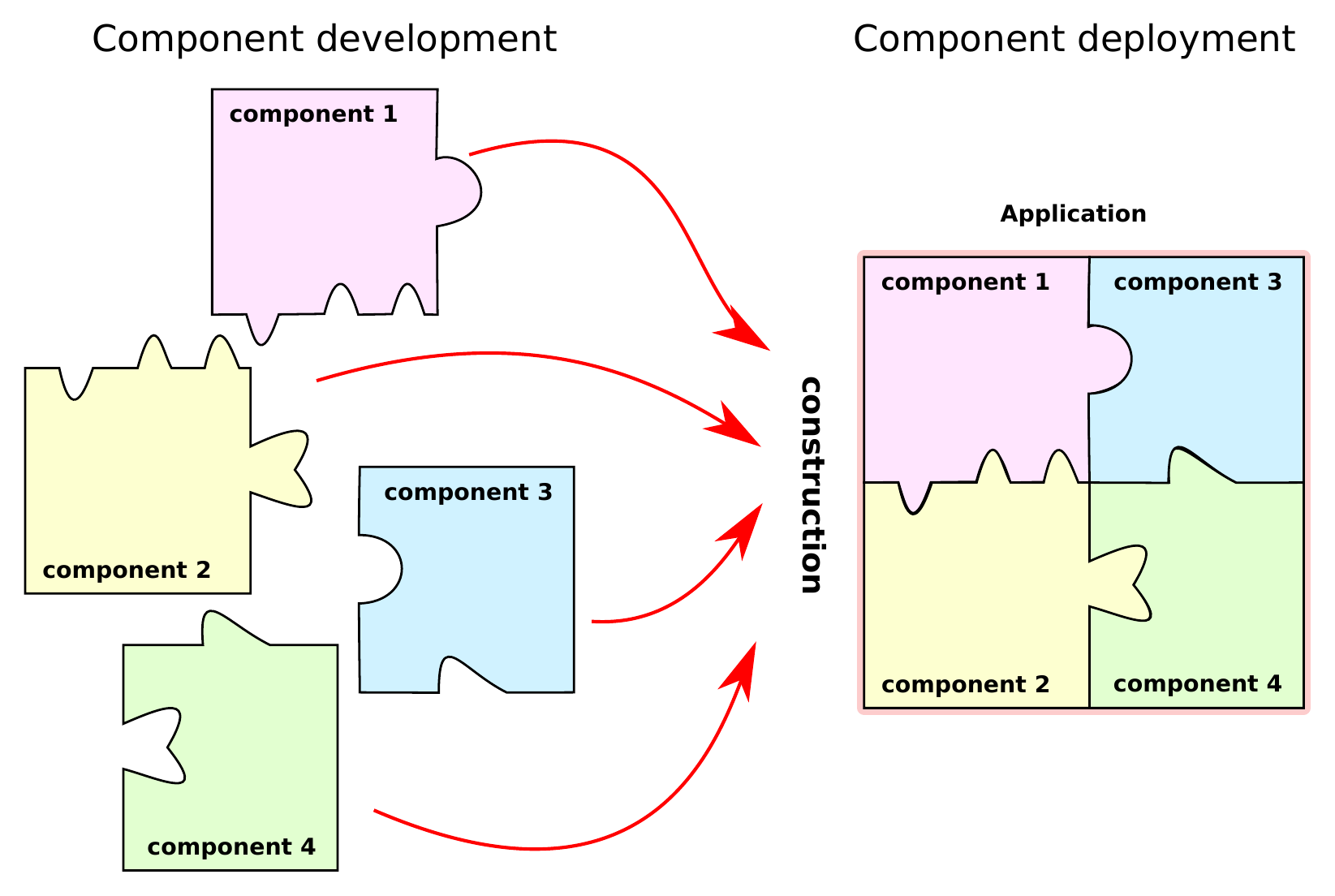}
   \caption{Component-based software architecture. The core idea is to separate the application into reusable components. In a mobile robot framework, path planning, for example, can be seen as one component that can be used in many applications and also on different hardware. Hardware-specific components, such as an interface to a specific sensor, allow the component to be reused within all systems that employ the particular sensor. Some components, for example, data logging, should be generic enough to only have one implementation for all different deployments. }
   \label{fig:component_architecture}
\end{figure}

Historically speaking, there has been a push in the robotics community toward universal software architectures. Over time, some research groups and research communities have invested more time in devising generic middleware systems. Frameworks that have offered more open interfaces, greater ease of software development and extension, more integrated functionality, and out of the box code for robot navigation, visualization, and simulation have naturally seen wider adoption in the community. Currently, there are several software architectures which have become very popular and are now considered standard tools.

A robotics framework is defined to be the complete package containing the full stack of software components (see Fig.~\ref{fig:component_architecture}) which are useful for developing software for a robotic system. In addition to the core communication tools (middleware), this commonly includes components for simulation, visualization, logging, and replaying data.

The key attractive feature of a robotic middleware is that it brings different components together and offers easy means to enable said components to communicate and interoperate. Another important aspect is that middleware also provides interfaces to sensors and actuators, which in many cases requires low-level access and therefore OS-specific calls to the system. A robotic middleware should assist the development of a complex robotic system both to handle the requirements in an adequate manner and to provide a well-structured API to functionality at different levels of interaction. For example, depending on the application at hand, the user may want to control precisely how the communication flow works, whereas in other cases a simple high-level API call would be sufficient.

\begin{figure}[ht!]
   \centering
   \includegraphics[width=\columnwidth]{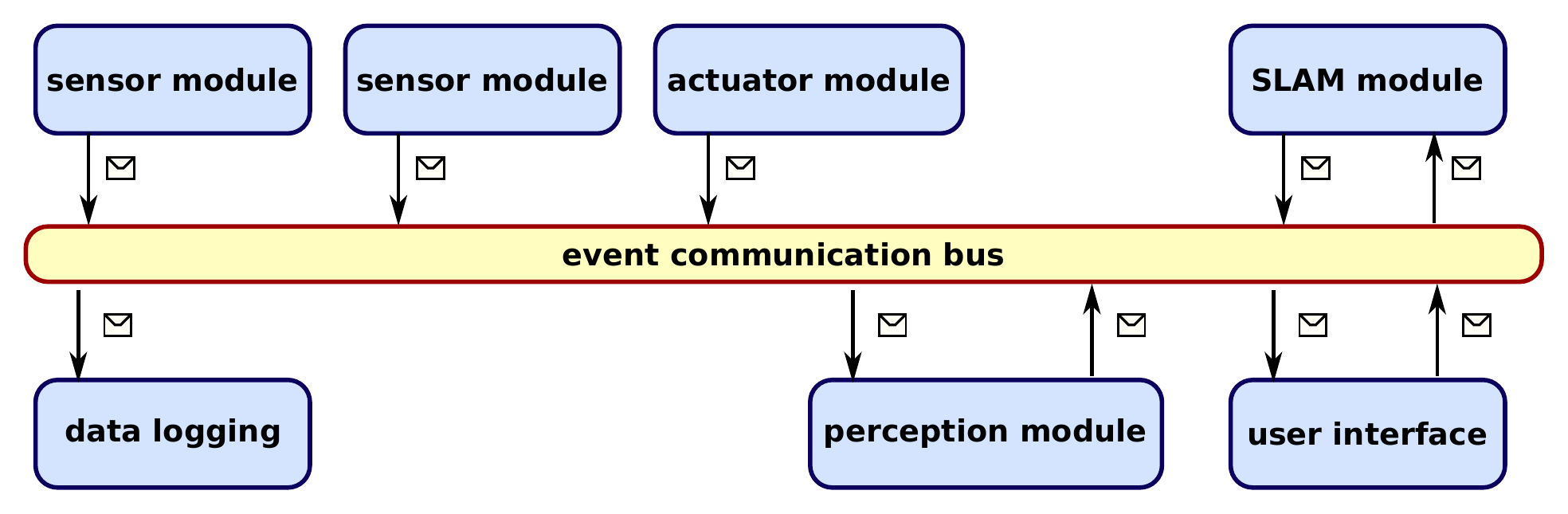}
   \caption{A loosely coupled design allows components to interact with each other in a flexible way and is provided through the middleware layer typically using a message-based event-driven architecture. Decoupling components allows them to start up, shut down, and re-initialize independently and in any given order: that is, in an ideal case, well-designed loosely coupled components do not interfere with each other.}
   \label{fig:loosely_coupled_design}
\end{figure}

As most frameworks use a modular design featuring a loosely coupled design pattern (see Fig.~\ref{fig:loosely_coupled_design}), the software architecture used within each component is less of an interest compared to how the infrastructure is designed to handle and set up communication between each component.
In essence, in this chapter we will look at software architectures that are commonly utilized in mobile robotics frameworks and especially pertaining to the design of the middleware.

\subsection{Robotic Frameworks}


\begin{table}
\caption{A set of commonly used and influential mobile robotic frameworks}
\label{tab:frameworks}
\scriptsize
\begin{tabular}{p{3.4cm}p{1cm}p{6cm}p{1cm}}
\hline\noalign{\smallskip}
Name & Active & Key Features & Open Source\\
\noalign{\smallskip}\svhline\noalign{\smallskip}
ROS -~\cite{ROS2009} & \cmark & Concurrent and asynchronous communication through messages, a central server is used to handle node communication. & \cmark\\
ROS2 -~\cite{Thomas2014} & \cmark & Concurrent and asynchronous communication through messages, decentralized handling of node communication. & \cmark\\
YARP -~\cite{YARP2006} & \cmark & Loosely coupled design, large support of various communication types. & \cmark\\
Orocos -~\cite{bruyninckx2001open} & \cmark & Real-time control of robots and machine tools, component-based and can be used in combination with ROS. & \cmark\\
Player -~\cite{Gerkey01mostvaluable} & \xmark & Multiple concurrent drivers communicating using a client/server model. & \cmark\\
Carmen -~\cite{Montemerlo03perspectiveson} & \xmark & Modular design, relies on IPC to handle message communication between isolated software executables. & \cmark\\
\noalign{\smallskip}\svhline\noalign{\smallskip}
MATLAB & \cmark & Utilize the ROS framework to interface sensors and actuators, large variety of generic toolkits that are useful in mobile robotics. & \xmark \\ 
MRDS & \xmark & Concurrent execution and message-based communication between isolated components. A central system is responsible for maintaining the interaction and running components. & \xmark \\
\noalign{\smallskip}\hline\noalign{\smallskip}
\end{tabular}
\vspace*{-12pt}
\end{table}

A vast number of different frameworks for mobile robotics are actively developed and maintained, both by the open source community, as well as by the commercial sector.
Some of the most influential and commonly used frameworks are listed in Table~\ref{tab:frameworks} and will be covered in more details below, but please note that this list is by no means complete, and many other frameworks exist, for example: OpenRDK~\citep{calisi2008openrdk}, OpenRTM~\citep{ando2008software}, PEIS-ecology~\citep{PEIS}, MIRO~\citep{utz2002miro}, OpenRave~\citep{diankov2008openrave}, Saphirai~\citep{konolige1997saphira}, MRPT~\citep{blanco2009phd} to mention a few.
The key aspect here is to draw general conclusions on the software architecture used rather than to detail out specific features of each framework.
For more details, we refer to several review articles overviewing a more comprehensive list of specific robotics middlewares~\citep{Elkady,Inigo-Blasco:2012:RSF:2184918.2184973,harris2011survey}.
Questions such as which requirements were met by different middlewares are of interest from an architectural design point of view.
Another interesting aspect is why some of the frameworks continued to develop and grow whereas others didn't.



\textbf{ROS - Robot Operating System} : The most used and widely spread mobile robotics middleware. Has a large active community and a wide support for various sensors, actuators, and mobile robots. Many companies support ROS by providing and maintaining interfaces to their products. Modular design with separate execution and data management, encapsulated into \textit{nodes}. Communication is performed using message passing over TCP (nodes) or shared memory (\textit{nodelets}). The system provides both a publisher/subscriber paradigm (see Fig.~\ref{fig:publish_subscribe}) for asynchronous communication and a client/server paradigm to enable synchronized execution on desired input data. A central server maintains discovery and connection handling between nodes, while the actual node connections are implemented in a peer-to-peer fashion. ROS supports a large variety of programming languages and includes both the communication interface infrastructure and a software packaging, building, and release toolchains. \\

\textbf{ROS2 - Robot Operating System 2} : The successor of ROS, where the key difference lies in the communication. ROS utilizes a custom-designed communication via TCP/IP. ROS2 instead utilizes DDS (\textit{Data Distribution Service}), which is a standardized way that has been used in a plethora of different systems and applications. 
DDS is the mechanism that handles the \textit{discovery} between \textit{nodes} in ROS2, which allows for a fully decentralized system. 
In ROS the mechanism of handling node communication is centralized and takes place through the \textit{roscore} software which acts as a \textit{master} where \textit{nodes} can ask the \textit{master} to \textit{discover} other nodes for them. In ROS2 this discovery mechanism is instead embedded in each node (via DDS), and there is no \textit{master} used in ROS2. This is, for example, helpful in multi-robot setups. Apart from these underlying changes, the overall functionality of ROS and ROS2 is similar. To transfer from ROS to ROS2 requires that the underlying code that builds up the \textit{nodes} has to be modified. Overall, ROS2 is gaining momentum but many packages are still only available for ROS. \\

\textbf{YARP - Yet Another Robot Platform} : Modular design, multi-threaded, multiple machines using a loosely coupled design pattern with a set of executables. It supports a large variety of connection types and a large variety of operating systems. YARP utilizes \textit{ports} in order to publish messages, similar to the usage of topics in ROS. A central name server, the \textit{yarpserver}, is typically used to connect software modules through their ports. In YARP any data type can be sent on the ports which in some cases is practical, but could also add more dependencies between components, since the data structure has to be shared. \\

\textbf{OROCOS - Open RObot Control Software} : Contains a set of different libraries, the core of which is the ``Real-Time Toolkit''. The ``Kinematics and Dynamics Library'' (KDL) is now integrated into the ROS framework and is an essential tool in the robot manipulator community. The design choice of the system is component-based, and the overarching aim is to obtain modules which are generic with respect to hardware and easy to integrate with other software packages. The key focus of the framework is real-time control of robots and machine tools. \\

\textbf{Player} : Was one of the most common middlewares to interface robots and senors in the early 2000s. It supports multiple concurrent connections using a client-server architecture. At the time, a large variety of hardware were supported. One key reason for its popularity was the usage of plug-in drivers which allowed drivers to be loaded as shared objects at runtime, thus separating driver interface code from the high-level core functionality. In addition to allowing encapsulation of driver code, this modular approach also enables easy sharing of code and is another important reason behind the popularity of Player. A key limitation of the framework is that it only supports a client-to-server setup and not point-to-point communication, which contributed to a general trend of migrating code out of Player and into alternative middleware frameworks, such as ROS. Along with Player two simulators were developed: Stage (2D) and Gazebo (3D); both these simulators have been integrated into other robotic frameworks. \\

\textbf{Carmen - Carnegie Mellon Robot Navigation Toolkit} : Modular structure. Communication between processes is done by passing messages using Inter-Process Communication (IPC). The framework features both publisher/subscriber pattern and client/server communication. It supports various platforms and features a navigation stack based on 2D laser scanner data. It contains a 2D simulator, supports message recording and playback, and utilizes a centralized parameter server. Messages are defined at compile time. \\

\textbf{MATLAB } : MATLAB comes with a large set of useful tools that could be utilized in a mobile robot setup. One toolkit, the Robotics System Toolbox, provides an interface between MATLAB and ROS. Essentially, MATLAB enables support to create nodes used in a ROS framework and is therefore not per se a framework on its own, but is one of the few commercial providers of software for mobile robotics.\\

\textbf{MRSD - Microsoft Robotics Developer Studio} : First released in 2006 and ended in 2014. As a framework it relies on an asynchronous library called CCR (Concurrency and Coordination Runtime) that utilizes messages to handle communication between isolated pieces of software called components (similar to nodes in the ROS world). Handling of components and their interactions is accomplished through DSS (Decentralized Software Services). A loosely coupling design pattern is utilized where each component has its own data management and each execution is done in isolation from the other components. It contains a full framework including a physic engine-based simulation package.\\


\section{Requirements of a Mobile Robotic Framework}

The requirements for a mobile robot stem from the requirements for the complete mobile robot system application. One difficulty here lies in the fact that there is a plethora of different mobile robot applications, hardware that needs to be interfaced, and desired system aspects to be optimized (e.g., positioning accuracy, real-time capabilities, and computational resources). A requirement is addressed within the middleware alone or by the complete framework. What really defines a hard requirement, and not merely a good feature, also depends on the application. A set of potential requirements is listed in Table~\ref{tab:requirements}, though this list is not exhaustive in any way, but merely illustrates what types of functionality are commonly needed. Notably, however, limitations that stem from faults in the middleware, are naturally of bigger concern, as they may be very hard or even impossible to address without rebuilding or restructuring the most fundamental parts of the system. Entries in the table directly addressable at framework level (labeled with the letter F) are in principle straightforward to incorporate and can be considered as nice to have features, rather than a hard requirement. 

\begin{table}
\caption{Common requirements along with solutions for a mobile robot system and at what level they are mainly addressed  (M - middleware, F - framework) }
\label{tab:requirements}
\scriptsize
\begin{tabular}{p{4.0cm}p{6.4cm}p{1cm}}
\hline\noalign{\smallskip}
Requirement & Common solution & Level\\
\noalign{\smallskip}\svhline\noalign{\smallskip}
``Soft'' real-time capabilities & Event-based triggering mechanism and a message-driven data flow & M \\
Distribute computation across multiple platforms & Component and message-based architecture & M \\
Robustness toward communication dropouts & Discover service and a hybrid peer-to-peer system & M \\
Reconfiguration of data flow at runtime & Loosely coupled component with a publisher/subscriber pattern & M \\
Efficient bandwidth usage & Peer-to-peer communication & M \\
Fault tolerance & Modular design, decoupling between components and self-sustained modules & M/F \\
Easy to develop and share software & Well documented API along with interface to native package management of the host OS & M/F \\
Easy to share and reuse code & Modular design, decoupling of components with clear interfaces, decoupling design & M/F \\
Support for multiple robots & Hybrid peer-to-peer design & M/F \\
Limit complexity of the software, e.g. dependencies & Component-based design with generic message types & M/F \\
Wide support of sensors and actuators & Large community, connects with easy to share and reuse code & F \\
Wide support of existing methods and algorithms & Large community, connect with easy to share and reuse code & F \\
Simulation/visualization support & A dedicated visualization/simulation software is included as a component into the framework which utilizes the same set of interfaces & F \\ 
Logging support & Separate software component that can publish or receive data on available interfaces & F \\ 

\noalign{\smallskip}\hline\noalign{\smallskip}
\end{tabular}
\vspace*{-12pt}
\end{table}





\begin{figure}[htb]
   \includegraphics[width=\linewidth]{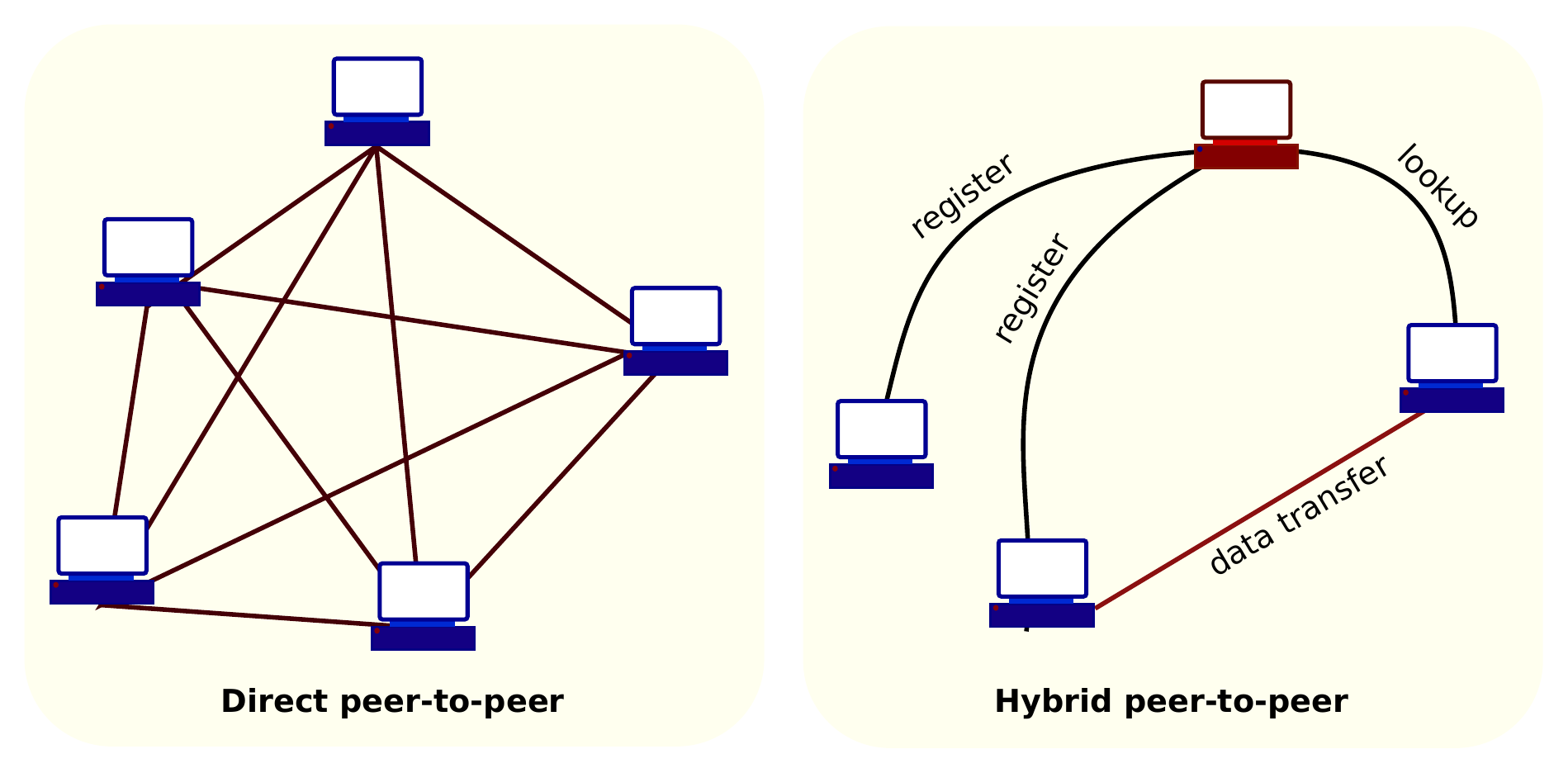}
   
   \caption{Plain peer-to-peer communication (left) and hybrid peer-to-peer communication (right). The core part of the hybrid system is to assist the peer's in finding each other and establishing the connection. }
   \label{fig:networks}
\end{figure}

From Table~\ref{tab:requirements} a few key features in the middleware stick out. One such feature is the need to have a concurrent and distributed system. In a robotic system, events occur in an unsynchronized manner, the different functionalities require different amounts of data, and the real-time aspect varies tremendously from high-level cycles (such as task planning) down to low-level control of individual actuators. All these functionalities have to interact and communicate with each other in a common framework. The current best practice is to avoid that all data is passed through a central server or hub, but instead enable each module to talk directly to any other module in the system, with as little intervention and overhead as possible. Here the current practice is to use a hybrid peer-to-peer architecture, where peer-to-peer denotes the direct communication between modules and where the hybrid part assists in establishing connections, see Fig.~\ref{fig:networks}.

Modular design refers to a design pattern whereby each functionality is isolated into a separate executable. The motivation is that if one part of the system fails, the whole system will not necessarily go down. With a properly decoupled setup, it should then be possible to restart the faulty component, without altering any of the other running components. Naturally, when having separated the system into modules, they have to communicate with each other in an efficient way. 
Splitting up the functionality in smaller components will invariably increase the amount of communication needed, and there is obviously a trade-off point where modularization makes the system more complex and cumbersome, removing any potential benefits due to communication overhead.
To allow each component to be independent, the interfaces exposed to other components are often implemented using hardware and software independent structures provided by the middleware. Due to the overhead involved in parsing and passing messages, especially for components that utilize large amount of data, shared memory can be utilized instead of a transmission protocol (called nodelets in the ROS world). On the downside, this approach limits how the components are distributed.

Mobile robots also come with a plethora of different onboard hardware components, ranging from small microcontroller-based system, to high-end computational devices, to ``off board'' computers and cloud services. Nowadays the hardware abstraction layer (HAL) is commonly an operating system, although microcontrollers are integrated by interfacing via CAN, Ethernet, serial bus to a PC or by running a lightweight OS supporting the framework at hand.

\section{Software Architecture for Mobile Robots}

From the requirement section above, there are a number of key design criteria that the robotic framework and the middleware should fulfill. Which common software architecture is suitable to address these requirements and which frameworks employ the different architectural patterns\footnote{\url{https://en.wikipedia.org/wiki/Software\_architecture}}?

\subsection{Blackboard}
One fundamental problem if multiple components or robots are used is how to share and allow concurrent access to information and modules.
One implementation is ``tuple spaces'' which follows a blackboard design.
The key purpose is that different components can be combined to solve a particular problem.
This approach has been implemented in the PEIS-Ecology framework~\cite{PEIS} which combined mobile robotics with ambient living.

Distributed shared memory is somewhat similar where a single memory can be shared among multiple components which in the context of mobile robots can be used to handle and update various parameters, which in ROS and CARMEN, for example, are denoted as the ``parameter server''.

\subsection{Client-Server Model}
A client-server model is widely used to separate different tasks across multiple machines, where a server typically serves multiple clients and the communication is initialized by the client.
The client-server model is utilized in some form in most frameworks but is too limiting to be used as the sole design pattern for a distributed system comprising a fleet of robots, which is one limitation of the Player framework.
Please note that many tasks are very suitable to have centralized processing, such as coordination and task allocation of multiple vehicles, and therefore it is vital that a mobile robotic framework supports it.
For example, in the ROS framework, the client-server model is used in a ``rosservice''.

\subsection{Component-Based Software Engineering}
A key concept utilized in most mobile robot software architectures is modularisation: i.e., splitting functionality into separate loosely coupled components.
Modules are directly interchangeable, given that they expose the same interfaces and, of course, provide the same functionality.
For example, a module is denoted ``node/device/driver'' in ROS/YARP/Player respectively.
In mobile robotic applications, there is a large variety of algorithms utilized to perform the same task.
For example, a path planner would as an input require information about the mobile robot, such as the shape and kinematic configuration.
Furthermore, it would require a map and a start and goal pose and will then output the generated path consisting of a set of robot poses, possibly with additional information such as curvature. By using the same input and output, an RRT- or lattice-based motion planner could easily be exchanged without the need to modify any other component.

The component-based design is a key factor of allowing code to be shared and reused. The flexibility to generate the required data structure used for passing data between nodes is also of importance. Here, for example, ROS uses messages with native data structures, such as ints and floats which can be defined along with the code of the component.

\subsection{Database-Centric}
A database-centric architecture builds around fetching and storing data in a shared container. Typically, a database system can be used as a communication layer between different processes, which is somewhat similar to the distributed memory described above.

Looking at mobile robot applications, there exist two types of data: online generated data, and post-processed/offline data.
Online data are generated from sensors while the robot is operating. 
For example, while building a map the representation of the environment is continuously updated using sensor data, where the size of the processed data is typically several magnitudes smaller than that of the combined raw data.
Processed data or offline data is commonly loaded during start-up, for example, a map (utilized for localization) or various lookup tables for path planning.
In general, offline or processed data comes with its own structure and with its own file format.
Using a database to store offline data is possible, but seldomly done in practice, with some notable exceptions such as YARP, which natively supports the Mongo database system.

The usage of databases to share online data with parallel processes in distributed systems is difficult, as it is at odds with the design choice of having a peer-to-peer-based communication semantic. Database storage of online data can also be inefficient, as in many applications the online data is processed once and never accessed again (e.g., as in robot localization).
Distributed database systems exist, but due to the cost of a larger overhead, more complex setup, and complication in fault handling, they are rarely used in practice in robotics frameworks.

\subsection{Event-Driven Architecture}
\begin{figure}[htb]
  \centering
  \includegraphics[width=\linewidth]{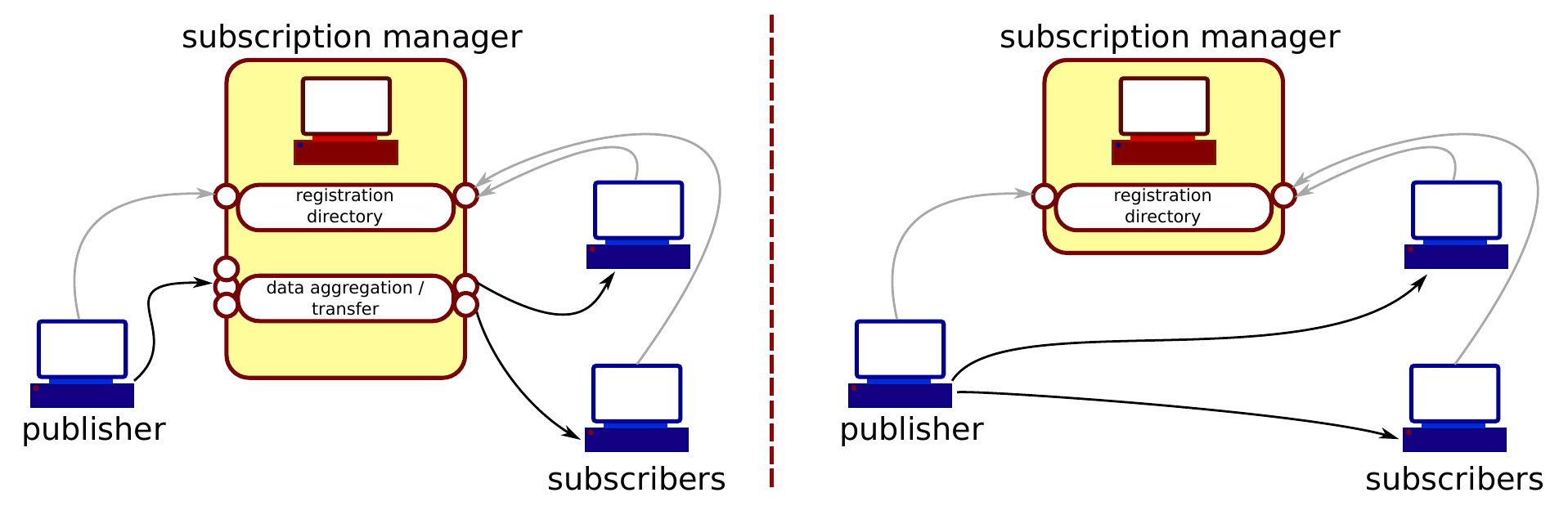}
   \caption{Publisher-subscriber model commonly used to decouple components. Both centralized (left), as well as hybrid peer-to-peer (right) implementations exist.}
   \label{fig:publish_subscribe}
\end{figure}

To use events to detect changes in states or occurrence of new data is common in mobile robotics middleware.
The event-driven mechanism is achieved through a message-driven architecture, and events are triggered when a message is passed around.
Message-driven architectures such as publisher/subscriber (Fig.~\ref{fig:publish_subscribe}) or client/server architecture utilize the concepts of events to trigger callbacks.
For example, when a new odometry measure is generated, that triggers an update in the current pose estimate and is typically used to propagate further messages around the system.
It is also common to utilize event messages to alter the behavior of a service-oriented architecture.
For example, a service providing paths to the robot could listen for events regarding occupancy of objects in the vicinity while generating motions. In essence all robotic middlewares employ an event-driven mechanism.

\subsection{Peer-to-Peer}
This architecture allows components to share information and tasks between components without the requirement of having a centralized server: a capability which is particularly important for multi-robot systems, but also has some desirable features even for single-robot architectures.
In the context of a multi-robot system, the functionality provided by each peer (robot) is typically rather similar, and many tasks can be equally well carried out by each peer.
For a mobile robot system, the peer network is typically configured in a hybrid fashion, by combining both peer-to-peer and client-server models.
In many multi-robot applications, there is a need for a central system, a single component/node, that takes care of dispatching robots or other high-level tasks such as coordination. Furthermore, having a central server is also beneficial to assist the peer-to-peer network in order to establish and find connections between peers.
This architectural design is commonly utilized in many robotic frameworks, such as ROS, YARP, CARMEN, etc. 

\subsection{Plug-In}
A plug-in is typically used to extend the functionality of an existing software component, without the need to change or re-compile the software. 
Plug-in components are implemented by utilizing common interfaces and shared libraries that can be loaded on demand.
For example, the visualization tool RViz in ROS utilizes this design in order to be able to visualize topics containing user-defined message types.
Note that plug-in capabilities are a general software engineering paradigm that can be implemented in any component, such as for example through ``pluginlib'' which extends plug-in functionalities to any ROS package.
The Gazebo simulator utilizes a plug-in structure in order for the user to add custom-defined sensors, actuators, and visualizations components. 
Another example is the plug-in architecture that Player uses in order to load its components called ``drivers'' as shared libraries at runtime.

\subsection{Publish-Subscribe Pattern}

A publisher/subscriber design allows for a loose coupling between a node which generates data and corresponding set of nodes which consume data. 
The design is usually implemented through a dedicated message handler commonly called a ``message broker''. 
The routing is done by the broker mechanism which allows on demand subscription and un-subscription of messages on different ports (YARP) or topics (ROS). 
A key difference compared to a client/server setup is that the client can only publish messages as long as the server is started, and vice versa.

\subsection{Service-Oriented Architecture: SOA}
A service-oriented architecture is related to component-based design, wherein each service provides a part of the solution and fits well in the peer-to-peer framework.
Compared to sensory data-driven tasks, such as localization which typically outputs a new estimate as soon as new messages arrive, a service can be seen as a single query potentially solving a larger problem.
The operational semantics also demands that it is crucial that the service call is processed, but the response time can be quite large, e.g., path planning.
For example, ROS denotes these types of requests as ``services'' and also supports ``action'' requests which allows the server side to return progress to the caller which is a useful interface for tasks which could take long time to complete.
``Actions'' could as well be implemented using a combination of service calls and a publisher/subscriber setup for the progress update.









\section{Future Direction for Research}

The software architecture for mobile robots has converged to rely on a hybrid peer-to-peer-based message parsing solution.
Messages are used within a publisher/subscriber paradigm to handle data in real time and in a client/server paradigm to provide solutions to non-real-time critical queries with non-deterministic execution times.
The currently most common mobile robotic frameworks today, YARP and ROS, have a similar structure.
It is also worth to note that the number of maintained framework reference implementations has significantly reduced during the past years and that many frameworks, including YARP, nowadays also interface ROS.

In the past the trend has been to go from a single monolithic system to more distributed component-based systems, and future work is envisioned to progress along the same direction: to enable better support and robustness for a more on-demand adjusted and efficient peer-to-peer setup. Currently the native support for having a dynamic load balancing across a set of multiple platforms within YARP and ROS is limited. The discovery service used to find new nodes is partially supported using separate packages in ROS and is now fully addressed in ROS2. 
Dynamic adjustment and balancing of bandwidth usage is not supported and is practically often even handled by setting up different network infrastructures and separating control commands from sensor streams to assure latency and throughput.

\bibliographystyle{spbasic}
\bibliography{sa4mr_ref}

\end{document}